# Vertical stratification of forest canopy for segmentation of under-story trees within small-footprint airborne LiDAR point clouds

Hamid Hamraz[a*], Marco A. Contreras[b], Jun Zhang[a]

a: Department of Computer Science, b: Department of Forestry

University of Kentucky, Lexington, KY 40506, USA

hhamraz@cs.uky.edu, marco.contreras@uky.edu, jzhang@cs.uky.edu

\* Corresponding Author:     hhamraz@cs.uky.edu  +1 (859) 489 1261

**Abstract**

Airborne LiDAR point cloud representing a forest contains 3D data, from which vertical stand structure even of under-story layers can be derived.  This paper presents a tree segmentation approach for multi-story stands that stratifies the point cloud to canopy layers and segments individual tree crowns within each layer using a digital surface model based tree segmentation method.  The novelty of the approach is the stratification procedure that separates the point cloud to an over-story and multiple under-story tree canopy layers by analyzing vertical distributions of LiDAR points within overlapping locales.  Unlike previous work that stripped stiff layers within a constrained area, the procedure stratifies the point cloud to flexible tree canopy layers over an unconstrained area with minimal over/under-segmentations of tree crowns across the layers.  The procedure does not make a priori assumptions about the shape and size of the tree crowns and can, independent of the tree segmentation method, be utilized to vertically stratify tree crowns of forest canopies with a variety of stand structures.  We applied the proposed approach to the University of Kentucky Robinson Forest – a natural deciduous forest with complex terrain and vegetation structure.  The segmentation results showed that using the stratification procedure strongly improved detecting under-story trees (from 46% to 68%) at the cost of introducing a fair number of over-segmented under-story trees (increased from 1% to 16%), while barely affecting the segmentation quality of over-story trees.  Results of vertical stratification of canopy showed that the point density of under-story canopy layers were suboptimal for performing reasonable tree segmentation, suggesting that acquiring denser LiDAR point clouds (becoming affordable due to advancements of the sensor technology and platforms) would allow more improvements in segmenting under-story trees.

**Keywords:** remote sensing, discrete return LiDAR, multi-story stand, canopy layering, individual tree segmentation.

# 1 Introduction

In the past two decades, airborne light detection and ranging (LiDAR) technology has extensively been used for forestry purposes due to its ability to capture data at unprecedented spatial and temporal resolutions in the shape of 3D point clouds (Ackermann 1999; Hyyppä et al. 2012; Maltamo et al. 2014; Swatantran et al. 2016; Wehr and Lohr 1999).  From this data, more detailed tree level information can be retrieved to improve the accuracy of forest assessment, monitoring, and management activities (Duncanson et al. 2012; Vastaranta et al. 2011; Weinacker et al. 2004; Wulder et al. 2012).   Due to the ability to penetrate vegetation canopy, LiDAR 3D point clouds also contain vertical information from which vegetation structural information even from under-story canopy layers can be retrieved (Hall et al. 2011; Lefsky et al. 2002; Maguya et al. 2014; Reutebuch et al. 2005), which is of great value for various forestry applications and ecological studies (Espírito-Santo et al. 2014; Ishii et al. 2004; Singh et al. 2015; Wing et al. 2012).  .  Although understory trees provide limited financial value and a minor proportion of total above ground biomass, they influence canopy succession and stand development, form a heterogeneous and dynamic habitat for numerous wildlife species, and are an essential component of forest ecosystems (Antos 2009; Heurich 2008; Jules et al. 2008; Moore et al. 2007).  However, to obtain individual trees attributes (e.g., location, crown width, height, DBH, volume, biomass) from different canopy layers, accurate and automated tree segmentation approaches that are able to separate tree crowns both vertically and horizontally are required (Duncanson et al. 2014; Ferraz et al. 2012; Shao and Reynolds 2006; Wang et al. 2008).

Numerous methods for individual tree segmentation within LiDAR data have been developed. Earlier methods use pre-processed data in the form of digital surface models (DSMs) or canopy height models (CHMs) to segment individual trees (Chen et al. 2006; Jing et al. 2012; Koch et al. 2006; Kwak et al. 2007; Popescu and Wynne 2004; Véga and Durrieu 2011).  These methods have an inherent drawback of missing under-story trees by considering only the surface data (Hamraz et al. 2016; Wang et al. 2008).  More recent methods process the raw point clouds in order to utilize all horizontal and vertical information and, from the computational viewpoint, can be classified to volumetric or profiler methods.  Volumetric methods directly search the 3D volume for the individual trees (Amiri et al. 2016; Ferraz et al. 2012; Lahivaara et al. 2014; Li et al. 2012; Lindberg et al. 2014; Lu et al. 2014; Rahman and Gorte 2009; Véga et al. 2014), hence are generally computationally intensive and may be prone to suboptimal solutions due to the large magnitude of the search space. On the other hand, profiler methods tame the computational load through a more modular process. They typically have a module for vertical segmentation, i.e., to strip the 3D volume to multiple 2D horizontal profiles, a module for horizontal segmentation, i.e., to search the trees within the profiles, and a module to ultimately aggregating the results across the profiles (Ayrey et al. 2017). However, they generally lose information about the vertical crown geometry when processing a 2D profile. To minimize information loss due to profiling, other profiler methods have analyzed vertical distribution of LiDAR points to identify 2.5D profiles embodying more information about vertical crown geometry. Wang et al. (2008) searched trees within each profile and used a top-down routine to unify any detected crown that may be present in different profiles. They analyzed vertical distribution of all LiDAR points globally within a given area to determine the height levels for stripping profiles. However, depending on the vegetation height variability, a globally derived height level may lead to under/over-segmenting tree crowns across the profiles. Other approaches addressed this issue by identifying constrained regions including one or more trees using a preliminary segmentation routine and independently 2.5D profiling each region (Duncanson et al. 2014; Paris et al. 2016; Popescu and Zhao 2008), yet the final result is dependent on the preliminary segmentation.

Although a number of methods for segmenting individual trees in multi-story stands have been proposed, they are still unable to satisfactorily detect most of the under-story trees. Typically, detection rate of dominant and co-dominant (over-story) trees is around or above 90% and detection rate of intermediate and overtopped (under-story) trees is below 50%. This inefficacy can be attributed to the reduced amount of LiDAR points penetrating below the main cohort formed by over-story trees (Kükenbrink et al. 2016; Takahashi et al. 2006), although incompetency of the current approaches to effectively use all vertical and horizontal information also plays a role. In this paper, we propose a profiler approach for segmenting crowns of all size trees in multi-story stands. The approach derives height levels locally hence stratifies the point cloud to 2.5D profiles (hereafter referred to as canopy layers), each of which is sensitive to stand height variability and includes a layer of non-overtopping tree crowns within an unconstrained area. The approach then utilizes a DSM-based method as a building block to segment individual tree crowns within each canopy layer.

## 2 Materials and Methods

### 2.1 Study site and LiDAR campaign

The study site is the University of Kentucky's Robinson Forest (RF, Lat. 37.4611, Long. -83.1555) located in the rugged eastern section of the Cumberland Plateau region of southeastern Kentucky in Breathitt, Perry, and Knott counties (Figure 1). The terrain across RF is characterized by a branching drainage pattern, creating narrow ridges with sandstone and siltstone rock formations, curving valleys and benched slopes. The slopes are dissected with many intermittent streams (Carpenter and Rumsey 1976) and are moderately steep ranging from 10 to over 100% facings predominately northwest and south east, and elevations ranging from 252 to 503 meters above sea level. Vegetation is composed of a diverse contiguous mixed mesophytic forest made up of approximately 80 tree species with northern red oak (*Quercus rubra*), white oak (*Quercus alba*), yellow-poplar (*Liriodendron tulipifera*), American beech (*Fagus grandifolia*), eastern hemlock (*Tsuga canadensis*) and sugar maple (*Acer saccharum*) as over-story species. Under-story species include eastern redbud (*Cercis canadensis*), flowering dogwood (*Cornus florida*), spicebush (*Lindera benzoin*), pawpaw (*Asimina triloba*), umbrella magnolia *(Magnolia tripetala)*, and bigleaf magnolia (*Magnolia macrophylla*) (Carpenter and Rumsey 1976; Overstreet 1984). Average canopy cover across RF is about 93% with small opening scattered throughout. Most areas exceed 97% canopy cover and recently harvested areas have an average cover as low as 63%. After being extensively logged in the 1920's, RF is considered second growth forest ranging from 80-100 years old, and is now protected from commercial logging and mining activities (Department of Forestry 2007). RF currently covers an aggregated area of ~7,440 ha, and includes about 2.5 million (±13.5%) trees of which over 60% are under-story (Hamraz et al. 2016, 2017b).

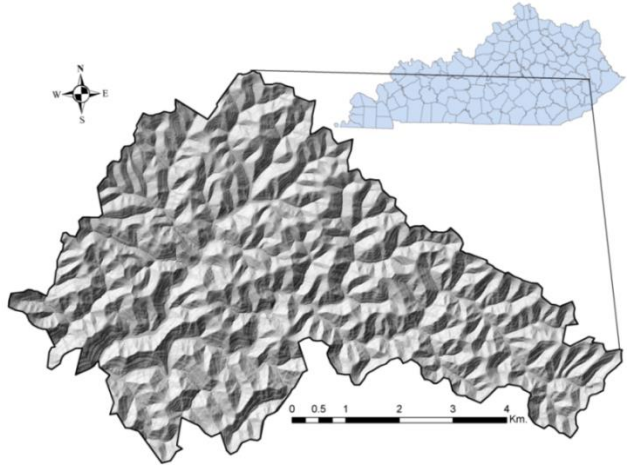

Figure 1. Terrain relief map of the University of Kentucky Robinson Forest and its general location within Kentucky, USA.

The LiDAR acquisition campaign over RF was performed in summer 2013 during leaf-on season (May 28-30) using a Leica ALS60 sensor, which was set at 40° field of view and 200 KHz pulse repetition rate. The sensor was flown at the average altitude of 200 m above ground at the speed of 105 knots with 50% swath overlap. Up to 4 returns were captured per pulse. Using the 95% middle portion of each swath, the resulting LiDAR dataset given the swath overlapping has an average density of 50 pt/m$^2$. The provider processed the raw LiDAR dataset using the TerraScan software (Terrasolid Ltd. 2012) to classify LiDAR points into ground and non-ground points. The ground points were then used to create a 1-meter resolution DEM using the natural neighbor as the fill void method and the average as the interpolation method.

## 2.2 Tree segmentation approach

Using the DEM, normalized heights of the LiDAR points are calculated then ground points are removed from further processing. The approach then stratifies the top canopy layer by analyzing the vertical distributions of the LiDAR points within overlapping locales and removes the layer from the point cloud. The approach then segments Individual tree crowns within the layer utilizing the DSM-based method introduced by Hamraz et al. (2016). Stratifying the top canopy layer of the remainder of the point cloud, removing it, and segmenting tree crowns within the layer iterates until the point cloud is emptied. Lastly, all tree crowns that have an average width of less than 1.5 m or are entirely located below 4 m from the ground (likely ground level vegetation) are removed as noise. Figure 2 visualizes the tree segmentation approach.

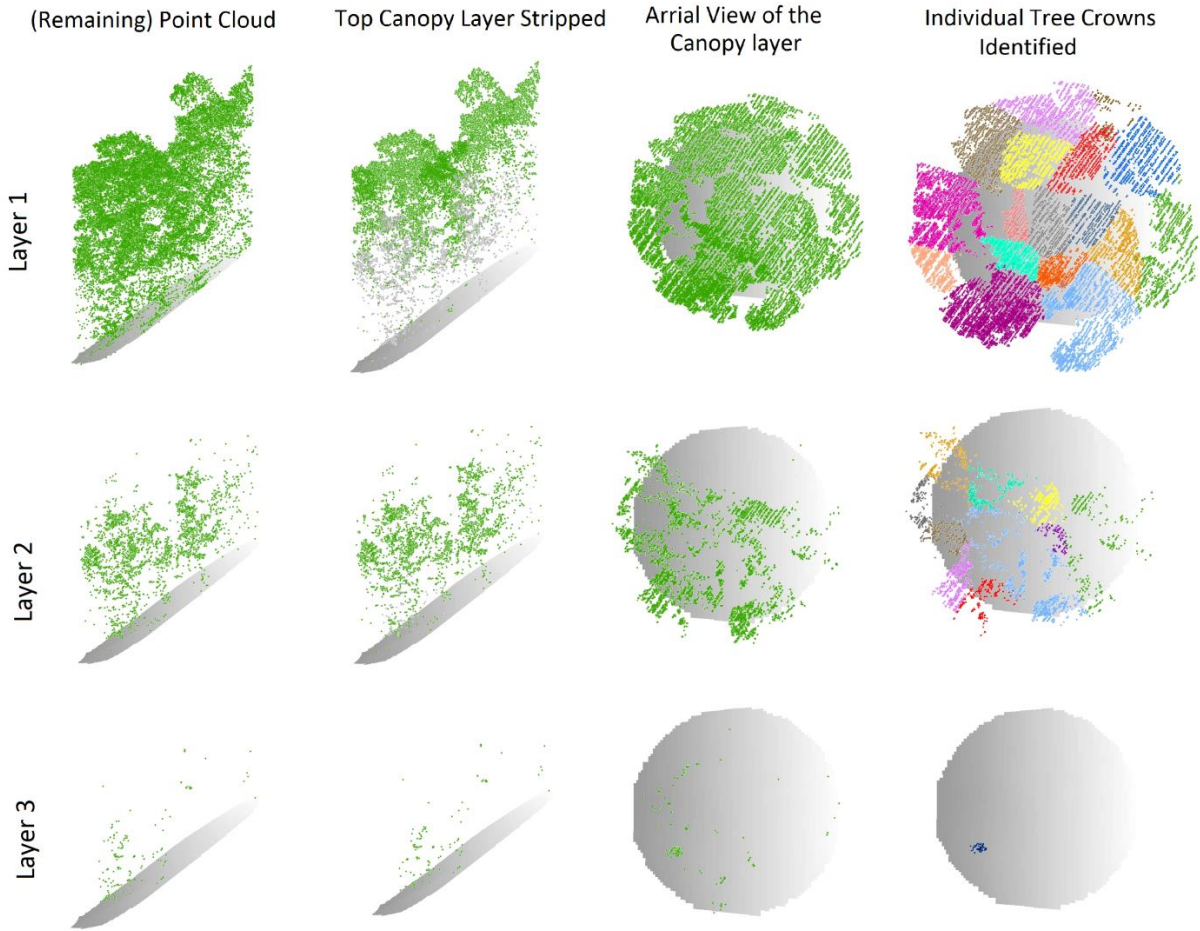

Figure 2. Illustration of the tree segmentation process in a multi-story stand by stratifying one canopy layer at a time, removing it from the point cloud, and segmenting crowns within it. A number of under-story trees seem to be missed within the third canopy layer, which is likely due to the much lower point density compared to the first and second layers.

To stratify the top canopy layer, the point cloud is binned into a horizontal grid with a cell width equal to the average footprint (AFP). AFP equals to the reciprocal of square root of point density, which itself is defined as the number of points divided by the horizontal area covered by the point cloud (as layers are removed from the point cloud, point density decreases hence AFP increases). The height threshold for removing the top layer is determined independently per each individual grid cell by inspecting the height histogram of all points in a circular locale around the cell, which should include sufficient number of points for building an empirical multi-modal distribution but not extending very far to preserve locality. We fixed the radius of the locale to 6×AFP (essentially containing about $\pi \times 6^2 > 100$ points), which is lower bounded at 1.5 m to prohibit too small locales capturing insufficient spatial structure.

To process a locale, we create a height histogram (bins fixed at 25 cm) of the points in the locale and smooth the histogram to remove variabilities pertaining to vertical structure of a single crown. We used a Gaussian filter with a standard deviation fixed at 5 m for smoothing. Every salient curve in the smoothed histogram, corresponding to a sequence of histogram bins throughout which the second derivative is negative, represents a canopy layer (Popescu and Zhao 2008; Wang et al. 2008), hence we chose the mid-point of the gap between the top and the second top layers as the height threshold for removing the top canopy layer within the cell location (Figure 3).

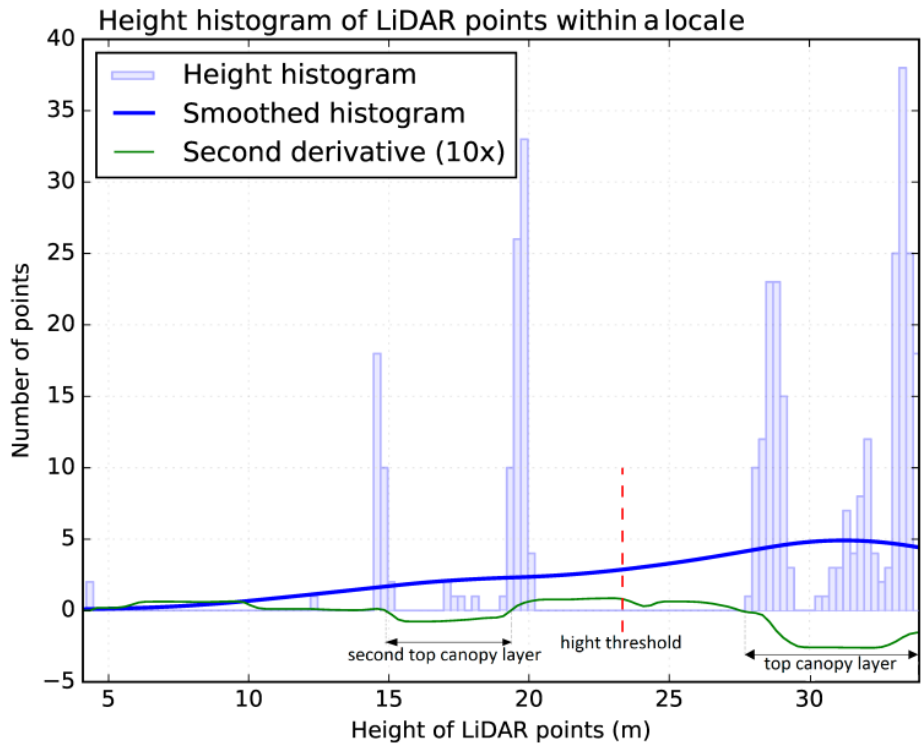

Figure 3. Height histogram of LiDAR points within a locale including over 100 points used for determining the height threshold for removing the top canopy layer in a cell location.

Since the height thresholds for removing the top canopy layer are determined using overlapping locales without a priori assumptions about tree crown shape or size, the canopy layer smoothly adjusts to incorporate vertical variabilities of crowns within an unconstrained area to minimize under/over-segmenting tree crowns (Figure 2), which is the major novelty of the proposed segmentation approach. Moreover, the canopy stratification procedure can be applied independent of the tree segmentation method in order to study the tree canopy layers of forested landscapes (Leiterer et al. 2015; Whitehurst et al. 2013).

## 2.3  Approach evaluation

### 2.3.1  Field data

Throughout the entire RF, 270 regularly distributed (grid-wise every 384 m) circular plots of 0.04 ha in size, centers of which were georeferenced with up to 5 m error,  were field surveyed during the summer of 2013.  Within each plot, DBH (cm), tree height (m), species, crown class (dominant, co-dominant, intermediate, overtopped), tree status (live, dead), and stem class (single, multiple) were recorded for all trees with DBH > than 12.5 cm.  In addition, horizontal distance and azimuth from plot center to the face of each tree at breast height were collected to create a stem map.  Site variables including slope, aspect, and slope position were also recorded for each plot.  Table 1 shows a summary of the plot level data.

Table 1. Summary of plot level data collected from the 270 plots in Robinson Forest.

| Plot-Level Metric | | Min | Max | Avg. | Total | Percent of total |
|---|---|---|---|---|---|---|
| Slope | (%) | 0 | 93 | 50 | | |
| Aspect | ⁰ | 2 | 360 | 179 | | |
| Tree count | | 2 | 41 | 14.7 | 3,971 | |
| Dominant | | 0 | 3 | 0.5 | 130 | 3.3 |
| Co-dominant | | 0 | 10 | 3.5 | 954 | 24.0 |
| Intermediate | | 0 | 34 | 5.5 | 1,481 | 37.3 |
| Overtopped | | 0 | 19 | 4.3 | 1,152 | 29.0 |
| Dead | | 0 | 7 | 0.9 | 254 | 6.4 |
| Species count | | 1 | 12 | 6.0 | 43 | |
| Shannon diversity index | | 0.0 | 2.25 | 1.50 | | |
| Average tree Height | (m) | 13.9 | 28.8 | 19.5 | | |
| Standard deviation of tree heights | (m) | 1.2 | 12.4 | 5.5 | | |

## 2.3.2 Evaluation method

LiDAR point clouds over each of the 270 field-surveyed plots included a 4.7-m buffer for capturing complete crowns of border trees using the proposed tree segmentation approach. The evaluation method assigns a score to each pair of LiDAR-derived tree location, assumed to be the apex of the segmented crown, and stem location measured in the field according to the tree height difference (should be less than 30%) and the leaning angle (should be less than 15° from nadir) between the crown apex and the stem location. It then selects the set of pairs with the maximum total score where each crown or stem location appears not more than once using the Hungarian assignment algorithm and regards the set as the matched trees (Hamraz et al. 2016; Kuhn 1955). The number of matched trees (MT) is an indication of the tree segmentation quality. The number of unmatched stem map locations (omission errors – OE) and unmatched LiDAR-derived crown apexes that are not in the buffer area (commission errors – CE) indicate under- and over-segmentation, respectively. The accuracy of the approach is calculated in terms of recall (Re – measure of tree detection rate), precision (Pr – measure of correctness of detected trees), and F-score (F – combined measure) using the following equations (Manning et al. 2008):

$$Re = \frac{MT}{MT + OE} \qquad (1)$$

$$Pr = \frac{MT}{MT + CE} \qquad (2)$$

$$F = 2 \times \frac{Re \times Pr}{Re + Pr} \qquad (3)$$

We evaluated the accuracy of the approach with and without canopy stratification (equivalent to the bare DSM-based method used in the approach) to assess the utility of the canopy stratification procedure for tree segmentation. We conducted two-tailed paired T-tests to compare the DSM-based and the stratification-enabled approach over nine accuracy metrics, i.e., precision, recall, and F-score for over-story, under-story, and all trees. We also present and discuss the summary metrics of the canopy layers stratified using the proposed procedure.

## 3 Results and Discussion

### 3.1 Tree segmentation accuracy

On average for the 270 sample plots, results from the DSM-based tree segmentation show higher precisions by 5–15% while the stratification-enabled approach shows higher recalls by 5–22% and higher F-scores by up to 12% (Figure 4). When comparing the stratification-enabled against the DSM-based approach using the T-tests (Table 2), all metrics except F-score for over-story trees showed significant (P < .0001) changes. Recall and precision for under-story trees showed the most remarkable changes: an increase of 22.1% (MSE = 10.035) and a decrease of 15.0% (MSE = 3.969), respectively. Overall, the stratification-enabled tree segmentation approach shows improvements in F-scores for under-story (by 11.52%, MSE = 1.698) as well as all trees (by 6.98%, MSE = 0.655), while barely affecting F-score for over-story trees compared with the DSM-based approach (Figure 4).

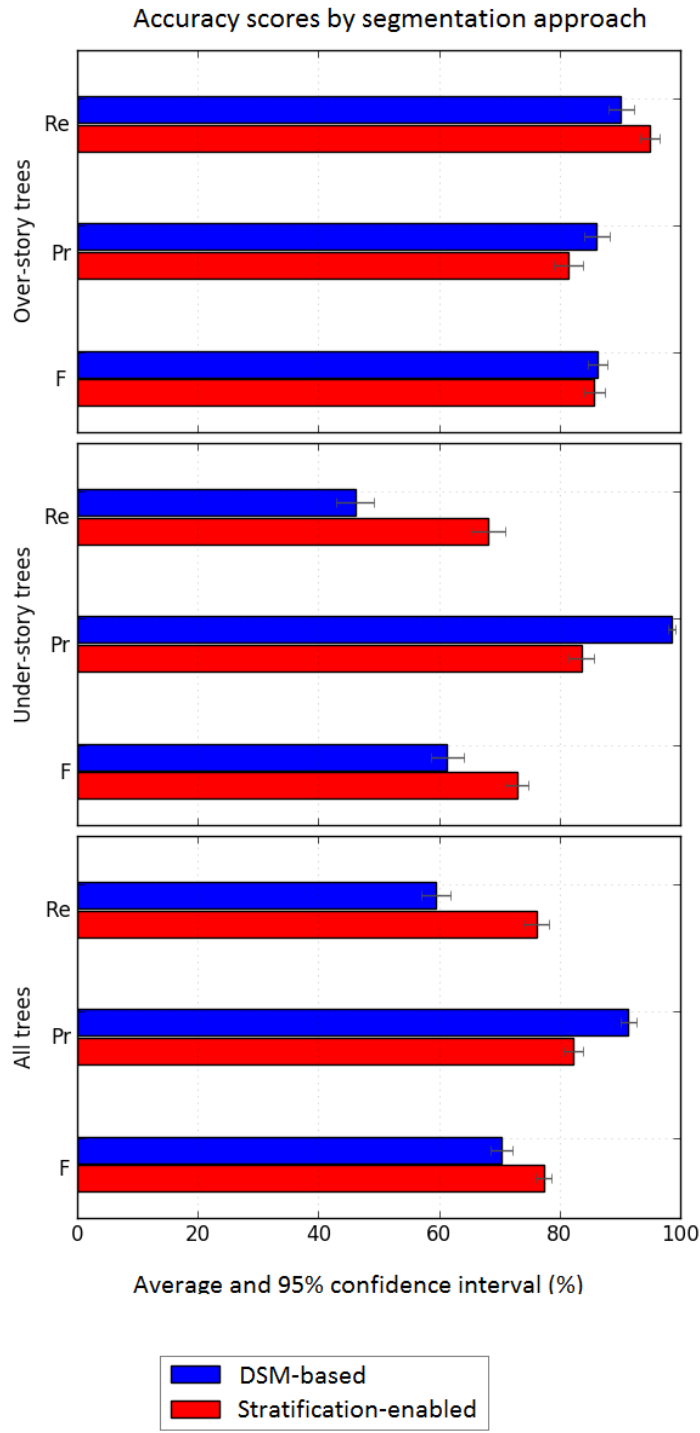

Figure 4. Average segmentation accuracies over 270 sample plots grouped by crown class.

Table 2. Summary of two-tailed paired T-tests assessing the improvement of canopy stratification for tree segmentation.

| Tree Class | Accuracy Metric | Samples Used | MSE | T-Score | P-Value | Average Improvement |
|---|---|---|---|---|---|---|
| Over-story | Re | 269 | 0.438 | 45.67 | <.0001 | +4.68% |
|  | Pr | 269 | 0.726 | 32.95 | <.0001 | -4.58% |
|  | F | 268 | 0.005 | 0.40 | 0.53 | -0.64% |
| Under-story | Re | 267 | 10.035 | 454.17 | <.0001 | +22.10% |
|  | Pr | 265 | 3.969 | 233.19 | <.0001 | -15.05% |
|  | F | 261 | 1.698 | 90.73 | <.0001 | +11.52% |
| All | Re | 270 | 5.440 | 473.70 | <.0001 | +16.56% |
|  | Pr | 270 | 1.744 | 175.00 | <.0001 | -8.98% |
|  | F | 269 | 0.655 | 76.39 | <.0001 | +6.98% |

Canopy stratification improved overall tree segmentation accuracy as benchmarked against a recently developed DSM-based segmentation method (Figure 4, Hamraz et al., 2016). However, this overall improvement is majorly composed of a strong increase in detection rate and a moderate decrease in correctness of the detected under-story trees. Detecting more trees likely increased the chance of over-segmentation of the detected trees, and this change was strongly pronounced for under-story trees compared with over-story ones. This observation indicates an increased sensitivity of the stratification-enabled approach to segment under-story trees while barely affecting the segmentation of over-story trees compared with the DSM-based method, which is also an indication of sound operation of the stratification procedure.

A few similar studies processed raw LiDAR point clouds and reported accuracy metrics for segmentation of under-story trees. For example, In a Norway spruce dominated forest, Solberg et al. (2006) detected 66% of the trees (dominant 93%, co-dominant 63%, intermediate 38%, and overtopped 19%) with a commission error of 26%. Paris et al. (2016) detected more than 90% of over-story and about 77% of under-story trees with a commission rate of 7% in conifer sites located in the Southern Italian Alps. However, due to tree crown architecture, segmenting trees in conifer stands is relatively simpler and studies have showed better performance compared to deciduous or mixed stands (Hu et al. 2014; Vauhkonen et al. 2011). In a deciduous stand at Smithsonian Environmental Research Center, Maryland, Duncanson et al. (2014) detected 70% of dominant (0% commissions), 58% of co-dominant (45% commissions), 35% of intermediate (166% commissions), and 21% of overtopped (29% commissions) trees. Ferraz et al. (2012) detected 99.3% of dominant, 92.6% of co-dominant, 65.7% of intermediate, and 14.5% of overtopped Eucalyptus trees in a Portuguese forest with an overall commission rate of 9.2%. In another deciduous stand in Eastern France, Véga et al. (2014) detected 100% and 44% of over-story and under-story trees with 27% and 3% commissions, respectively. Detection rate of our stratification-enabled tree segmentation approach was ~95% for over-story trees and ~68% for under-story trees, with a commission rate of ~17%. These results show improvements, especially in segmenting under-story trees, bearing the caveat that aforementioned studies were conducted in different sites using different LiDAR acquisition parameters with slightly different field surveying protocols and evaluation methods.

### 3.2 Canopy stratification procedure

For most of the 270 plots, the stratification procedure identified three (68.2%) or four (24.1%) canopy layers with an expected number of canopy layers of 3.16 per plot. Any layer located below 4 m for its entirety was excluded because it likely represents ground level vegetation, though any of the remaining layers may extend below 4 m and even touch the ground. Starting height and thickness of a canopy layer are defined as the medians over all grid cells used to stratify the layer (Figure 3). The average starting height of a canopy layer ranged between 0.3 to 18.2 m and the average thickness of a layer ranged between 6.1 and 8.8 m. Also, the average point density of a layer ranged between 0.44 and 42.08 $pt/m^2$. The average starting height, thickness, and point density of the entire canopy (all layers aggregated) was 1.4 m, 24.8 m, and 47.45 $pt/m^2$, respectively. The average point density of the entire canopy agrees with the average point density of the initial LiDAR dataset of 50 $pt/m^2$ given that ground and ground level vegetation returns were removed.

Table 3. Summary statistics of the canopy layers stratified within the 270 sample plots.

| Canopy Layer | Plots[1] | Starting Height (m) Avg. | S.D. | Thickness (m) Avg. | S.D. | Point Density (pt/m$^2$) Avg. | S.D. |
|---|---|---|---|---|---|---|---|
| 1 | 0.00% | 18.16 | 4.53 | 8.18 | 0.38 | 42.08 | 17.42 |
| 2 | 7.78% | 4.23 | 2.58 | 8.76 | 0.99 | 5.02 | 3.23 |
| 3 | 68.15% | 0.47 | 1.03 | 6.44 | 1.35 | 0.84 | 0.79 |
| 4 | 24.07% | 0.34 | 1.39 | 6.14 | 1.82 | 0.44 | 0.80 |
| Aggregate | 100.00% | 1.38 | 1.41 | 24.85 | 4.26 | 47.45 | 20.13 |

[1] Plots having as many number of canopy layers.

Thickness and point density generally decreases with lower canopy layers (Table 3). Specifically, the two lower canopy layers, where the majority of under-story trees are found, have an average density lower than 1 pt/m$^2$ (Table 3). Such low density is far less than the optimal point density (~4 pt/m$^2$) for segmenting individual trees (Evans et al. 2009; Jakubowski et al. 2013; Wallace et al. 2014), which is the main reason for inferior tree segmentation accuracy of under-story trees compared with over-story trees. Moreover, lower canopy layers are more tightly placed compared with higher canopy layers as also shown by Whitehurst et al. (2013), which might have made stratification of the layers more challenging and increased the chances of under/over-segmentation of small under-story trees.

As reported by Kükenbrink et al. (2016), at least 25% of canopy volume remain uncovered even in small-footprint airborne LiDAR acquisition campaigns, which concurs with suboptimal point density of lower canopy layers for tree segmentation in our study. If, however, our initial point cloud was a few times denser, the two lower canopy layers might have neared the optimal density, likely boosting segmentation accuracy of under-story trees. In a concurrent study, we modeled how point density of lower canopy layers decreases and estimated that a point cloud density of about 170 pt/m$^2$ is required to segment under-story trees within as deep as the third canopy layer with accuracies similar to over-story trees (Hamraz et al. 2017a). Such dense LiDAR campaigns are slowly becoming affordable given the advancements of the sensor technology and platforms (Swatantran et al. 2016).

Lastly, an interesting counter intuitive observation was that thickness of a canopy layer seemed to be unrelated to its starting height except only for very low starting heights (Figure 5), which is likely associated with layers formed by very small trees. Dependence of a canopy layer thickness on the number of layers preceding it and its independence to height is likely due to the fact that tree crowns within a canopy layer adapt their shape to maximize light exposure (Duursma and Mäkelä 2007; OSADA and TAKEDA 2003), and light exposure is related to the amount of light already intercepted by preceding canopy layers rather than the height of the layer.

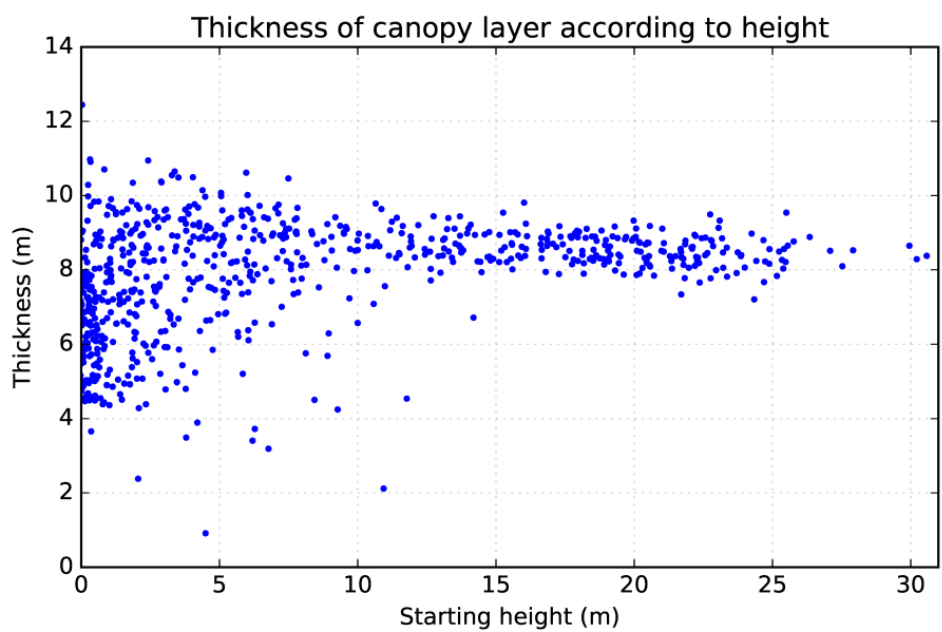

Figure 5. Thickness of canopy layer according to starting height of the layer.

## 4  Conclusions

Small-footprint LiDAR data covering forested areas contain a wealth of information of both horizontal and vertical vegetation structure that can be utilized to enhance various forestry applications and ecological studies. In this paper, we presented a modular approach that stratified the raw point cloud extended over an unconstrained area to its tree canopy layers, and utilized a DSM-based tree crown segmentation method as a building block for each layer to segment all sized trees in a multi-story deciduous stand.  Statistical analyses showed overall improvements in segmentation accuracy of under-story trees without any noticeable change in the accuracy of over-story trees, which was the main objective of using canopy stratification as a module for tree segmentation.  The proposed canopy stratification procedure can also be applied independent of the crown segmentation method in order to vertically stratify canopy to flexible layers of tree crowns over unconstrained areas.

The modular process of our segmentation approach allowed us to study the canopy layers individually.  We observed that the point densities of the lower canopy layers were suboptimal for segmentation of individual under-story trees.  It is expected that acquiring denser LiDAR point clouds brings the point density of lower canopy layers closer to optimal value, likely resulting in more improvements in the segmentation of under-story trees.  The result presented indicates this work is a promising step forward toward correctly retrieving and modeling all individual (over-story and under-story) trees of a natural forest using small-footprint LiDAR data.

## Acknowledgments

This work was supported by: 1) the Department of Forestry at the University of Kentucky and the McIntire-Stennis project KY009026 Accession 1001477, ii) the Kentucky Science and Engineering Foundation under the grant KSEF-3405-RDE-018, and iii) the University of Kentucky Center for Computational Sciences.  The authors would also like to thank Chase Clark for preparing the 3D visualization in Figure 2.

## Data availability

The datasets generated during and/or analysed during the current study are available from the corresponding author on reasonable request.

Amiri, N., Yao, W., Heurich, M., Krzystek, P., & Skidmore, A.K. (2016). Estimation of regeneration coverage in a temperate forest by 3D segmentation using airborne laser scanning data. *International Journal of Applied Earth Observation and Geoinformation, 52*, 252-262

Antos, J. (2009). Understory plants in temperate forests. *Forests and forest plants. Eolss Publishers Co Ltd, Oxford*, 262-279

Ayrey, E., Fraver, S., Kershaw Jr, J.A., Kenefic, L.S., Hayes, D., Weiskittel, A.R., & Roth, B.E. (2017). Layer Stacking: A Novel Algorithm for Individual Forest Tree Segmentation from LiDAR Point Clouds. *Canadian Journal of Remote Sensing*, 1-13

Carpenter, S.B., & Rumsey, R.L. (1976). Trees and shrubs of Robinson Forest Breathitt County, Kentucky. *Castanea*, 277-282

Chen, Q., Baldocchi, D., Gong, P., & Kelly, M. (2006). Isolating individual trees in a savanna woodland using small-footprint LiDAR data. *Photogrammetric Engineering and Remote Sensing, 72*, 923-932

Department of Forestry (2007). Robinson Forest: a facility for research, teaching, and extension education. In: University of Kentucky

Duncanson, L., Cook, B., Hurtt, G., & Dubayah, R. (2014). An efficient, multi-layered crown delineation algorithm for mapping individual tree structure across multiple ecosystems. *Remote Sensing of Environment, 154*, 378-386

Duncanson, L., Dubayah, R., Hurtt, G., Pinto, N., Cook, B., & Swatantran, A. (2012). How important is individual tree information for biomass modeling and mapping? In, *AGU Fall Meeting Abstracts* (p. 0353)

Duursma, R., & Mäkelä, A. (2007). Summary models for light interception and light-use efficiency of non-homogeneous canopies. *Tree physiology, 27*, 859-870

Espírito-Santo, F.D., Gloor, M., Keller, M., Malhi, Y., Saatchi, S., Nelson, B., Junior, R.C.O., Pereira, C., Lloyd, J., & Frolking, S. (2014). Size and frequency of natural forest disturbances and the Amazon forest carbon balance. *Nature communications, 5*

Evans, J.S., Hudak, A.T., Faux, R., & Smith, A. (2009). Discrete return lidar in natural resources: Recommendations for project planning, data processing, and deliverables. *Remote Sensing, 1*, 776-794
16

Véga, C., Hamrouni, A., El Mokhtari, S., Morel, J., Bock, J., Renaud, J.-P., Bouvier, M., & Durrieu, S. (2014). PTrees: A point-based approach to forest tree extraction from lidar data. *International Journal of Applied Earth Observation and Geoinformation, 33*, 98-108

Wallace, L., Lucieer, A., & Watson, C.S. (2014). Evaluating tree detection and segmentation routines on very high resolution UAV LiDAR data. *IEEE Transactions on Geoscience and Remote Sensing, 52*, 7619-7628

Wang, Y., Weinacker, H., & Koch, B. (2008). A lidar point cloud based procedure for vertical canopy structure analysis and 3D single tree modelling in forest. *Sensors, 8*, 3938-3951

Wehr, A., & Lohr, U. (1999). Airborne laser scanning—an introduction and overview. *ISPRS Journal of Photogrammetry and Remote Sensing, 54*, 68-82

Weinacker, H., Koch, B., Heyder, U., & Weinacker, R. (2004). Development of filtering, segmentation and modelling modules for lidar and multispectral data as a fundament of an automatic forest inventory system. *International Archives of Photogrammetry, Remote Sensing and Spatial Information Sciences, 36 (Part 8)*, W2

Whitehurst, A.S., Swatantran, A., Blair, J.B., Hofton, M.A., & Dubayah, R. (2013). Characterization of canopy layering in forested ecosystems using full waveform lidar. *Remote Sensing, 5*, 2014-2036

Wing, B.M., Ritchie, M.W., Boston, K., Cohen, W.B., Gitelman, A., & Olsen, M.J. (2012). Prediction of understory vegetation cover with airborne lidar in an interior ponderosa pine forest. *Remote Sensing of Environment, 124*, 730-741

Wulder, M.A., White, J.C., Nelson, R.F., Næsset, E., Ørka, H.O., Coops, N.C., Hilker, T., Bater, C.W., & Gobakken, T. (2012). Lidar sampling for large-area forest characterization: A review. *Remote Sensing of Environment, 121*, 196-209
21